\documentclass[sigconf,authorversion,nonacm]{acmart}

\usepackage{multirow}
\usepackage{subcaption}
\usepackage{makecell}

\begin{document}
\title{Using AntiPatterns to avoid MLOps Mistakes}

\author{Nikil Muralidhar\footnotemark[2], Sathappan Muthiah\footnotemark[2], Patrick Butler\footnotemark[2], \\ Manish Jain\footnotemark[4], Yu Yu\footnotemark[4], Katy Burne\footnotemark[4], Weipeng Li\footnotemark[4], David Jones\footnotemark[4]\\Prakash Arunachalam\footnotemark[4], Hays `Skip' McCormick\footnotemark[4], and Naren Ramakrishnan\footnotemark[2]\\
\footnotemark[2]Department of Computer Science, Virginia Tech, Arlington, VA 22203\\
\footnotemark[4]The Bank of New York Mellon, 240 Greenwich Street, New York, NY 10286}

\renewcommand{\shortauthors}{Muralidhar and Muthiah et al.}
\renewcommand{\shorttitle}{MLOps with Financial Applications}
\newcommand{\nikhilc}[1]{\textcolor{red}{#1}}

\begin{abstract}
We describe lessons learned from developing and deploying machine learning models at scale across the enterprise in a range of
financial analytics applications. These lessons are presented in the form of {\it antipatterns}.
Just as design patterns codify best software engineering practices, antipatterns provide a vocabulary to describe defective practices and methodologies. Here we catalog and document numerous antipatterns in financial ML  operations (MLOps).
Some antipatterns are due to technical errors, while others are due to not having sufficient knowledge of the surrounding context in which ML results are used. By providing a common vocabulary to discuss these situations, our intent is that antipatterns will support better documentation of issues, rapid communication between stakeholders, and faster resolution of problems. In addition
to cataloging antipatterns, we describe solutions, best practices, and future directions toward MLOps maturity.
\end{abstract}

\maketitle

\section{Introduction}
The runaway success of machine learning models has given rise to a better understanding of the technical challenges underlying their widespread deployment~\cite{paleyes2020challenges,sculley2015hidden}. There is now a
viewpoint~\cite{isbell2020} encouraging the rethinking of ML as a software
engineering enterprise.
MLOps---Machine Learning Operations---refers to the
body of work that focuses
on the full
lifecycle of ML model deployment, performance
tracking, and ensuring
stability in
production pipelines. 

At the Bank of New-York Mellon (a large-scale investment banking, custodial banking, and asset servicing enterprise) we have developed a range of enterprise-scale ML pipelines, spanning areas such as customer attrition forecasting, predicting
treasury settlement failures, and balance prediction. In deploying
these pipelines,
we have encountered several recurring antipatterns~\cite{antipatterns}
that we wish to document
in this paper.
Just as design patterns codify best software engineering practices, antipatterns provide a vocabulary to describe defective practices and methodologies.
Antipatterns often
turn out to be commonly utilized approaches that are actually bad,
in the sense that the consequences outweigh any benefits. Using antipatterns to desribe what is happening helps ML teams get past any blamestorming and arrive at a  refactored solution more quickly. While we do not provide a completed formal antipattern taxonomy,
our intent here is
to
support better documentation of issues, rapid communication
between stakeholders, and faster resolution of problems.

Our goals are similar to
the work of~\cite{sculley2015hidden} that argues for
the study of MLOps through
the lens of {\it hidden
technical debt.} While many of the lessons from~\cite{sculley2015hidden} dovetail with our
own conclusions, our
perspective here is
complementary, viz. we focus less on software engineering but more on data pipelines, how data is transduced into decisions, and how feedback from decisions can (and should) be
used to adjust and improve the ML pipeline.
In
particular, our study recognizes
the role of multiple
stakeholders (beyond ML
developers) who play crucial
roles in the success of
ML systems.

Our main contributions are:
\begin{enumerate}
    \item We provide a vocabulary of antipatterns that we have encountered in ML pipelines, especially in the financial analytics domain. While many appear obvious in retrospect we believe cataloging them here will contribute to greater understanding and maturity of ML pipelines.
    \item We argue for a new approach that rethinks ML deployment not just in terms of predictive performance but in terms of a multi-stage decision making loop involving humans. This leads to a more nuanced understanding of ML objectives and how evaluation criteria dovetail with deployment considerations.
    %\item We describe approaches to reduce unfairness and bias that we have found relevant in the enterprise.
    \item Finally, similar to Model Cards~\cite{modelcards}, we provide several recommendations for documenting and managing MLOps at an enterprise scale. In particular we describe the crucial role played by model certification authorities in the enterprise.
\end{enumerate}

\section{Case Study: Forecasting Treasury Fails}

\begin{figure*}
\begin{minipage}{0.4\textwidth}
\includegraphics[scale=0.2]{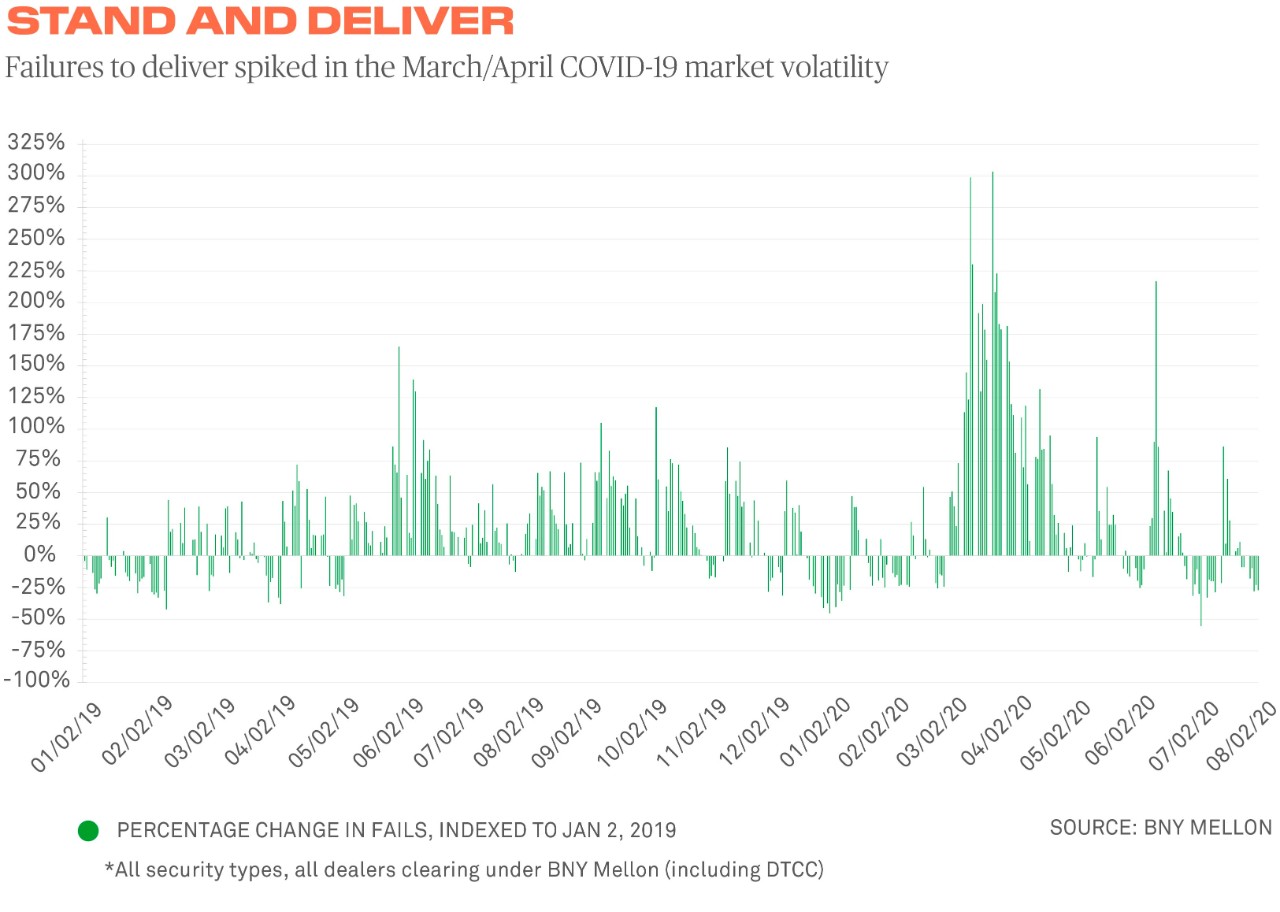}
\subcaption{}
\label{fig:percentage_change_fails}
\end{minipage}
\hfill
\begin{minipage}{0.45\textwidth}
\includegraphics[scale=0.28]{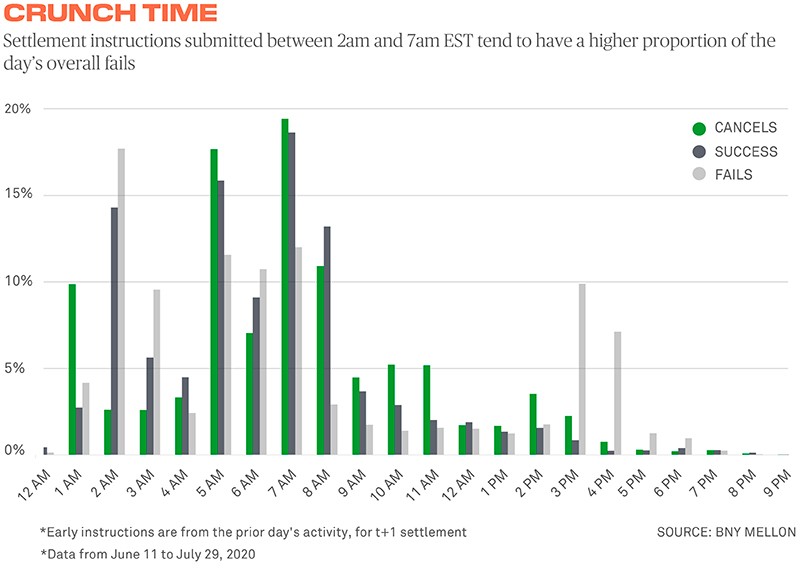}
\vspace{1.3em}
\subcaption{}
\label{fig:crunchtime}
\end{minipage}
\caption{Forecasting US Treasury Settlement Fails. (a) Heightened market volatility during March and April from COVID-19, when traders switched to working remotely, led to difficulties for firms in making sure everything was running smoothly. The larger volume of securities settlements in that period contributed to a higher number of fails. (b) Settlement instructions submitted between 2am and 7am NY time have a proportionally higher failure rate because the trade instructions are submitted with less visibility into the day’s market conditions. }
\label{fig:blog_fig1_fig2}
\end{figure*}

Capital markets have long suffered from a nagging problem: every day, roughly 2\% of all U.S. Treasuries and mortgage-backed securities set to change hands between buyers and sellers do not end up with their new owners by the time they are supposed to arrive.
Such `fails' happen for
many reasons, e.g.,
unique patterns in trading,
supply and demand imbalances,
speediness of given
securities, operational
hiccups, or credit events.
After the collapse
of Lehman Brothers, which led to an increase in settlement fails, the
Treasury Market Practices Group (TPMG) in our organization recommended daily
penalty charges on fails
to promote better market
functioning. 
The failed-to party
generally requests and
recoups the TPMG fails
charge from the non-delivering counterparty. After broad
industry adoption, according to the Federal
Reserve, the prevailing rate of settlement fails has fallen considerably.

In the middle of the COVID-19 market crisis, demand for cash and cash-like instruments such as Treasuries was drastically higher than normal, compounding the issue of settlement fails. Fig.~\ref{fig:percentage_change_fails} showcases the fallout of COVID-19 on the market in the form of settlement fails during March and April 2020. Liquidity issues in the Treasury market prompted the Fed to step in and buy more of the securities to restore calm.

We \iffalse BNY Mellon has\fi have developed a machine learning service that
uses intraday metrics
and other signals as early indicators of liquidity issues in specific sets of bonds to forecast
settlement
failures by 1:30pm daily
NY time.  The service also takes into account elements like the velocity of trading in a given security across different time horizons, the volume of bonds circulating, a bond’s scarcity, the number of trades settled every hour and any operational issues, such as higher-than-normal cancellation rates. Fig.~\ref{fig:crunchtime} showcases the daily failure rate dynamics (per hour) and characterizes the complexity of the task that the aforementioned machine learning service is modeling.
The resulting predictions help \iffalse BNY Mellon’s\fi our clients, including bond dealers, to monitor their intraday positions much more closely, manage down their liquidity buffers for more effective regulatory capital treatment, and offset their risks of failed settlements.
Through this and other ML services we have gained significant insight into MLOps issues that we aim to showcase here.

In developing and deploying this application, we encounter issues such as:
\begin{enumerate}
    \item Does the data processing pipeline have unintended side-effects on modeling due to data leakage or HARKing~\cite{gencoglu2019hark}? (Sections ~\ref{sec:data_leakage},~\ref{sec:testing_and_evaluation})
   \item What happens when models `misbehave' in production? How is this misbehavior measured? Are there compensatory or remedial pipelines? (Sections~\ref{sec:meta_modeling},~\ref{sec:calibration})
    \item How often are models re-trained and what is the process necessary to tune models? Is the training and model tuning reproducible? (Section ~\ref{sec:hyperparameter_tuning})
    \item How is model performance assessed and tracked to ensure compliance with performance requirements? (Sections ~\ref{sec:baselines},~\ref{sec:bad_credit_assignment})
    \item What constitutes a material change in the MLOps pipeline? How are changes handled? (Section~\ref{sec:concept_drift})
    \item Where does the input data reside and how is it prepared on a regular basis for input to an ML model? (Section~\ref{sec:data_crisis_as_a_service})
\end{enumerate}
\iffalse 
\begin{enumerate}
    \item Where does the input data reside and how is it prepared on a regular basis for input to an ML model? (Section~\ref{sec:data_crisis_as_a_service})
    \item How often are models re-trained and what is the process necessary to tune models? Is the training and model tuning reproducible? (Section ~\ref{sec:hyperparameter_tuning})
    \item How is model performance assessed and tracked to ensure compliance with performance requirements? (Sections ~\ref{sec:baselines},~\ref{sec:bad_credit_assignment})
    \item Does the data processing pipeline have unintended side-effects on modeling due to data leakage or HARKing~\cite{gencoglu2019hark}? (Sections ~\ref{sec:data_leakage},~\ref{sec:testing_and_evaluation})
    \item Who certifies models for production and how often?
    \item What happens when models `misbehave' in production? How is this misbehavior measured? Are there compensatory or remedial pipelines? (Sections~\ref{sec:meta_modeling},~\ref{sec:calibration})
    \item What constitutes a material change in the MLOps pipeline? How are changes handled? (Section~\ref{sec:concept_drift})
\end{enumerate}
\fi 
\noindent
Any organization employing ML in production needs to grapple with (at least) each of the questions above. In the process of doing so, they might encounter several
antipatterns as we document below.

%%%%%%% AntiPatterns Table 
\begin{table}[!ht]
\centering
\caption{Nine commonly practiced AntiPatterns.}
\begin{tabular}{c|l}
\toprule
\textbf{Stage} & \textbf{Name} \\ \toprule
\multirow{2}{*}{\textbf{\makecell{Design \& \\ Development}}} & Data Leakage\\ \cline{2-2}
\vspace{-0.2cm}
\\&  Tuning-under-the-Carpet \\ \midrule\midrule
\multirow{3}{*}{\hspace{-0.1cm}\textbf{\makecell{Performance Evaluation}}} &  `PEST' \\ \cline{2-2}
& Bad Credit Assignment \\ \cline{2-2}
& Grade-your-own-Exam \\ \midrule\midrule
\multirow{4}{*}{\textbf{\makecell{Deployment \& \\Maintenance}}} & `Act Now, Reflect Never' \\ \cline{2-2}
& Set \& Forget \\ \cline{2-2}
& `Communicate with Ambivalence'\\ \cline{2-2}
&  `Data Crisis as a Service'\\ \bottomrule
\end{tabular}
\label{tab:antipatterns_summary}
\end{table}

\section{AntiPatterns}
For the most part, we present our antipatterns (summarized in Table~\ref{tab:antipatterns_summary}) in a supervised learning or forecasting context. In a production ML context, there is typically a model that has been approved for daily use. Over time, such a model might be replaced by a newer (e.g., more accurate) model, or retrained with more recent data (but keeping existing hyperparameters or ranges constant or fixed), or retrained with new search for hyperparameters in addition to retraining with recent data. %These \emph{hyperparameters} have a significant effect of the representation learned by the model. 
In this process, we encounter a range of methodological issues leading to several antipatterns, which we identify below.

\begin{figure*}[!ht]
    \begin{minipage}{0.45\textwidth}
        \centering
        \includegraphics[scale=0.6]{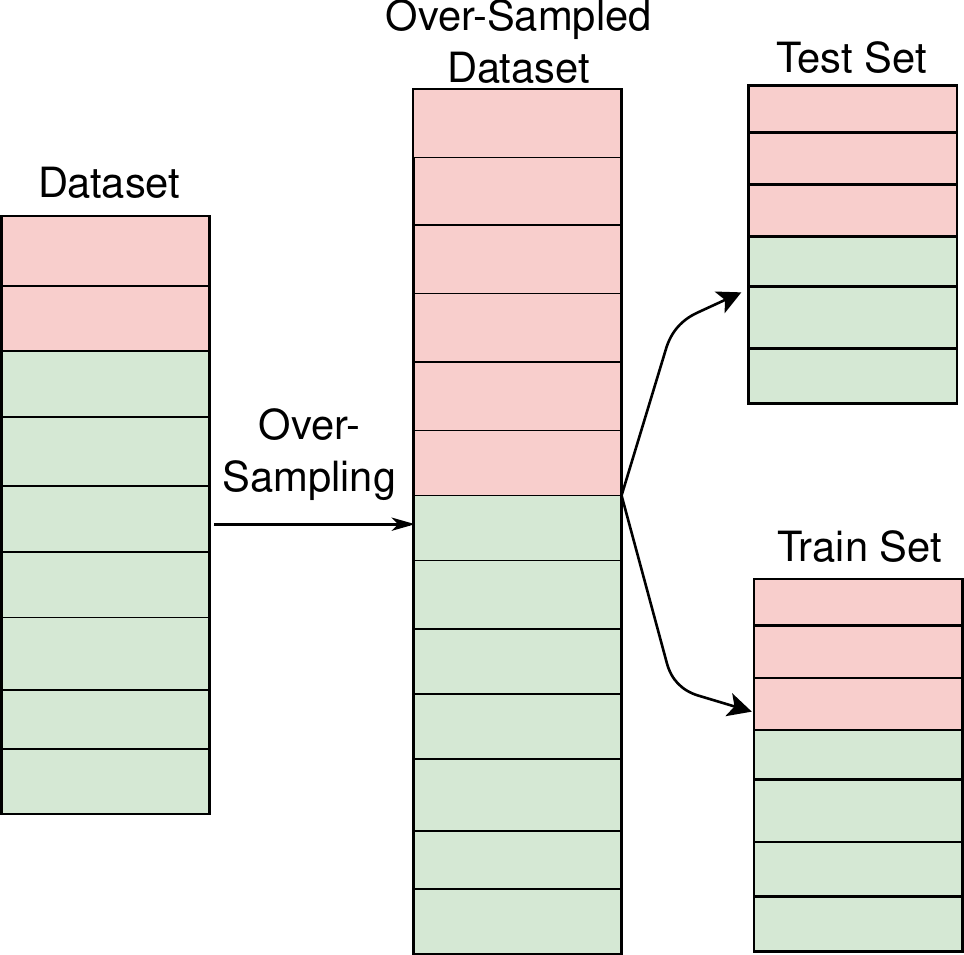}
        \label{fig:sampling_leakage1}
        \vspace{1cm}
    \end{minipage}
    \hfill
    \begin{minipage}{0.45\textwidth}
        \centering
        \includegraphics[scale=0.6]{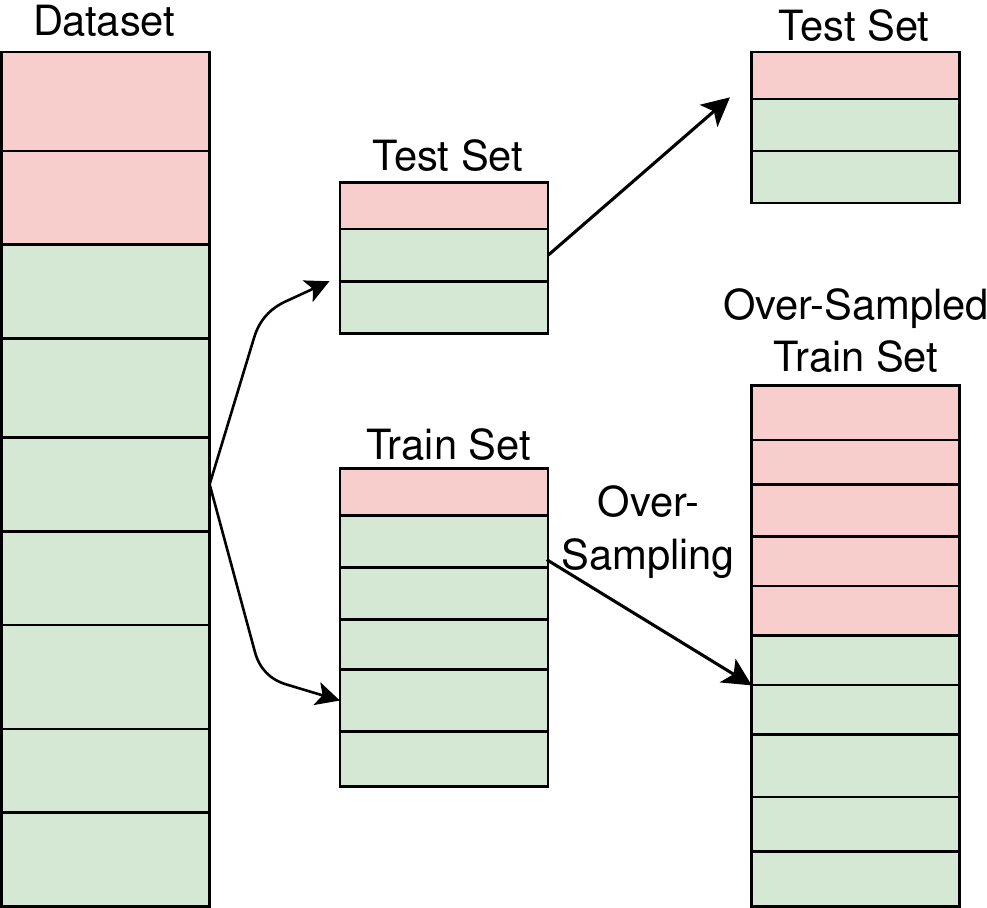}
        \label{fig:sampling_leakage2}
    \end{minipage}
    \vfill
    \begin{minipage}{0.45\textwidth}
        \centering
        \subcaption{Sampling Leakge}
        \begin{tabular}{lrrrr}
            \toprule
            {} &  Prec. &  Rec. &  F1 &   Sup. \\
            \midrule
            $\neg$ Attrited &      0.84 &   0.83 &     0.84 & 2,550.00 \\
            Attrited &      0.83 &   0.85 &     0.84 & 2,550.00 \\
            acc.          &      0.84 &   0.84 &     0.84 &    5,100.00 \\
            macro avg.         &      0.84 &   0.84 &     0.84 & 5,100.00 \\
            wt. avg.      &      0.84 &   0.84 &     0.84 & 5,100.00 \\
        \bottomrule
        \end{tabular}
        \label{tab:samp_leak}
    \end{minipage}
    \hfill
    \begin{minipage}{0.45\textwidth}
        \centering
        \subcaption{No Sampling Leakge}
        \begin{tabular}{lrrrr}
            \toprule
            {} &  Prec. &  Rec. &  F1 &   Sup. \\
            \midrule
            $\neg$ Attrited &      0.96 &   0.83 &     0.89 & 2,551.00 \\
            Attrited &      0.48 &   0.81 &     0.60 &   488.00 \\
            acc.          &      0.83 &   0.83 &     0.83 &   3,039.00 \\
            macro avg.         &      0.72 &   0.82 &     0.75 & 3,039.00 \\
            wt. avg.      &      0.88 &   0.83 &     0.84 & 3,039.00 \\
            \bottomrule
        \end{tabular}
        \label{tab:no_samp_leak}
    \end{minipage}
    \caption{Here we characterize the effect of oversampling in the financial application of banking customer churn (i.e., Attrition) detection. (a) Illustrates the pipeline wherein oversampling is carried out before separating the data for training and evaluation. (b) Illustrates the oversampling pipeline wherein the data for training and evaluation is first separated and only the training dataset is over-sampled. We can see that the pipeline in (a) shows better optimistic performance (i.e., F1 score for \emph{Attrited} class in Table~\ref{tab:samp_leak}) than (b) (i.e., F1 score for \emph{Attrited} class in Table~\ref{tab:no_samp_leak}) due to leakage in information from over-sampling before selecting the test set}
    \label{fig:sampling_leakage_plots}
\end{figure*}
\subsection{Data Leakage AntiPattern}\label{sec:data_leakage}
The separation of
training and test while
extolled in every ML101 course can sometimes be violated in insidious ways. Data leakage refers broadly to 
scenarios wherein
a model makes use of information that it is not supposed to have or would not have available in production. Data leakage leads to overly optimistic model performance estimates and poses serious downstream problems upon model deployment (specifically in high risk applications).
Leakage can happen sometimes unintentionally
when feature selection is driven by model validation or test performance or due to the presence of (typically unavailable) features highly correlated with the label. 
Samala et. al.~\cite{samala2020hazards} talk more about the hazards of leakage, paying
particular attention to
medical imaging applications. In our domain of financial analytics, 
increasingly complex features are constantly developed such that their complexity masks underlying temporal
dependencies which are
often the primary causes
of leakage. Below are
specific leakage
antipatterns we have encountered.

\subsubsection{\bf Peek-a-Boo AntiPattern}
Many source time-series datasets are based on reporting that lags the
actual measurement. A good example is Jobs data which is reported in the following month.
Modelers who are simply consuming this data may not be cognizant that the date of availability
lags the date of the data, and unwittingly include it in their models inappropriately.

\subsubsection{\bf Temporal Leakage AntiPattern.}
When constructing training and test datasets by sampling, the process
by which such sampling
is conducted can cause
leakage and thus lead to
not truly independent
training and test
sets.
In forecasting
problems, especially,
temporal leakage happens
when the training and
test split is not
carried out sequentially, thereby leading to high correlation (owing to temporal dependence and causality) between the two sets. 
    
\subsubsection{\bf Oversampling Leakage AntiPattern.}
An egregious form of leakage can be
termed \emph{oversampling leakage}, seen in situations involving a minority class. A well known strategy to use in imbalanced classification is to perform minority over-sampling, e.g., via an algorithm such as SMOTE~\cite{smote}. In such situations, if over-sampling is performed {\it before} splitting into training and test sets, then there is a possibility of information leakage. Due to the subtle nature of this type of leakage, we showcase illustrations and performance characterizations of oversampling leakage, in the context of customer churn detection in banking transactions in Fig.~\ref{fig:sampling_leakage_plots}. 

\subsubsection{\bf Metrics-from-Beyond AntiPattern.} 
This type of antipattern
can also be seen as
pre-processing leakage or
hyper-parameter leakage.
Often times, due to carelessness in pre-processing data, both training and test datasets are grouped and standardized together leading to leakage of test data statistics. For example when using standard normalization, if test and train datasets are normalized together, then the  sample mean and variance used for normalization is a function of the test set and thus
leakage has occurred.

\subsection{`Act Now, Reflect Never' AntiPattern}\label{sec:meta_modeling}
Once models are placed in production, we have seen that predictions are sometimes
used as-is without any filtering, updating, reflection, or even periodic manual inspection. This is an issue especially in situations where we see 1) concept drift (discussed in section~\ref{sec:concept_drift}), 2) irrelevant or easily recognisable erroneous predictions, and 3) adversarial attacks.

\begin{figure*}[!ht]
    \begin{minipage}{0.45\textwidth}
            \includegraphics[scale=0.2]{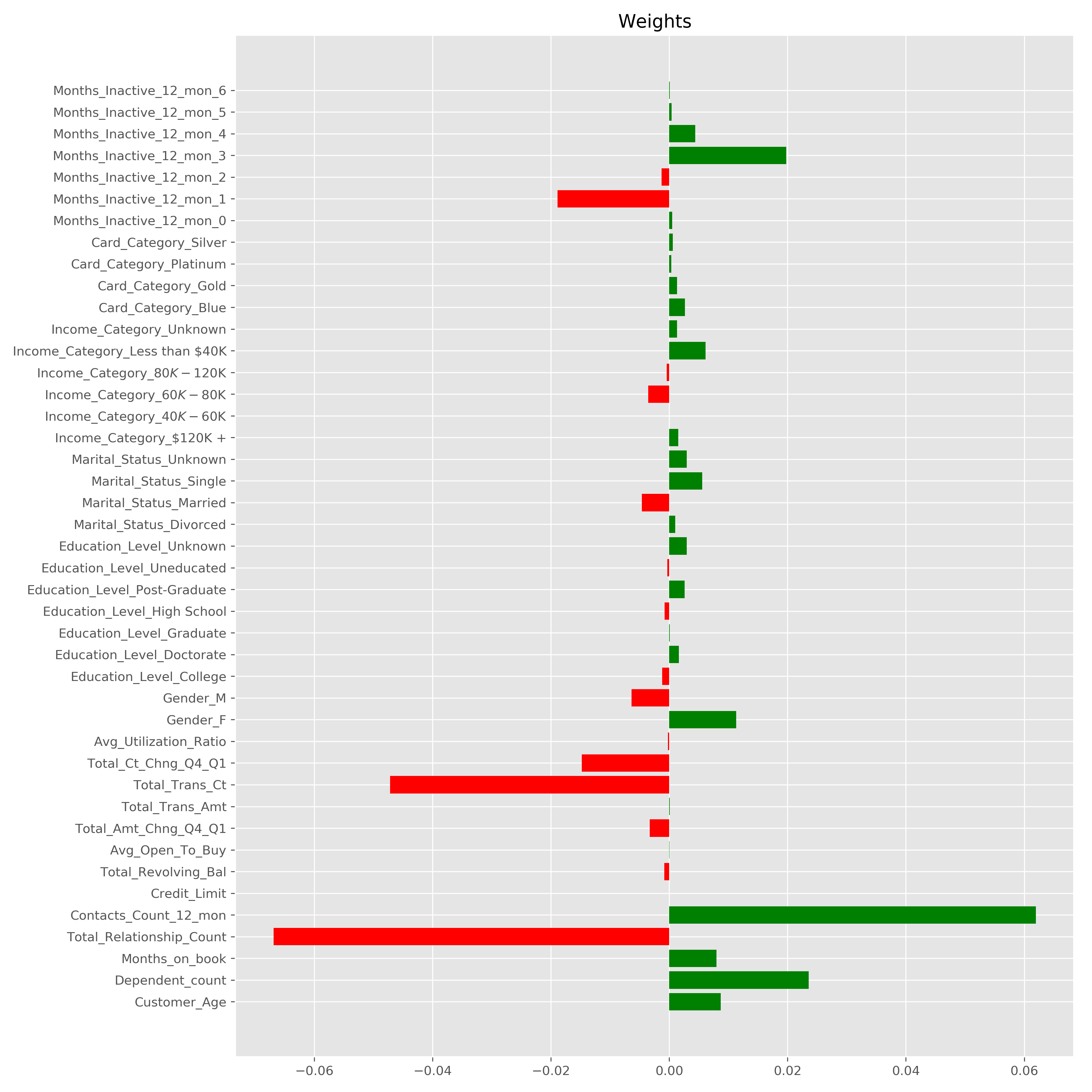}
            \subcaption{Feature Importance Characterization}
    \end{minipage}
    \hfill
    \begin{minipage}{0.45\textwidth}
            \includegraphics[scale=0.376]{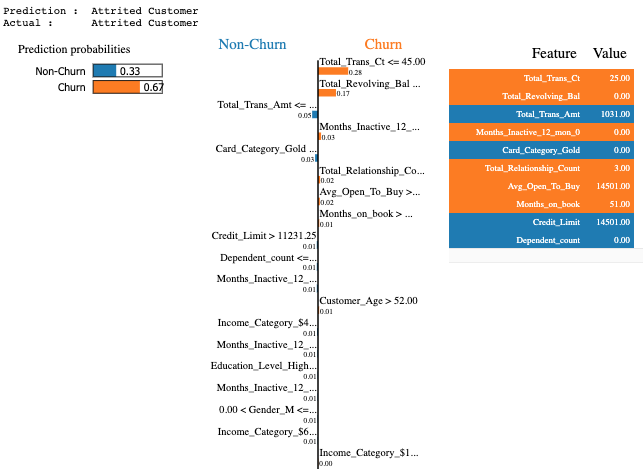}
            \subcaption{Single Instance Explanation using LIME framework}
    \end{minipage}
    \caption{Inspecting Model Decisions using explanation
    frameworks. Demonstrated on churn (i.e., Attrition) detection application using the LIME~\cite{ribeiro2016should} framework.}
    \label{fig:meta-modeling}
\end{figure*}

It is important to have systems in place that can monitor, track, and debug deployed models. For instance, under such situations it can be productive to have a meta-model that evaluates every model prediction and deems if it is trustworthy (or of required quality) to be delivered. For example, Ramakrishnan et. al.~\cite{beatingthenews} describe a meta-model called the fusion and suppression system that is responsible for the generation of final set of alerts from an underlying alert-stream originating from multiple ML models. The fusion and suppression system is responsible for performing duplicate detection, filling in missing values, and is also used to fine-tune precision / recall by suppressing alerts deemed to be of low quality.  A second solution could be to inspect model decisions further by employing explanation frameworks like LIME~\cite{ribeiro2016should}. Fig.~\ref{fig:meta-modeling} characterizes modeling decisions using meta-modeling frameworks.

\subsection{Tuning-under-the-Carpet AntiPattern}\label{sec:hyperparameter_tuning}
Different values of hyper-parameters often prove to be significant drivers of model performance and are expensive to tune and mostly task specific. Hyper-parameters play such a crucial role in modeling architectures that entire research efforts are devoted to developing efficient hyper-parameter search strategies~\cite{bergstra2013making,nguyen2019bayesian,henderson2018deep,van2018hyperparameter,probst2019tunability}.

The set of hyper-parameters differs for different learning algorithms. For instance, even a simple classification model like the decision tree classifier, has hyper-parameters like the maximum depth of the tree, the minimum number of samples to split an internal node and the criterion to use for estimating either the impurity at a node (gini) or the information gain (entropy) at each node. Ensemble models like random forest classifiers and gradient boosting machines also have additional parameters governing the number of estimators (trees) to include in the model. Another popular classifier, the support-vector machine which is a maximum margin classifier requires the specification of hyper-parameters that govern the type of kernel used (polynomial, radial-basis-function, linear etc.) as well as the penalty for mis-classification which in-turn governs the margin of the decision boundary learned. For an exhaustive analysis of the effect of hyper-parameters, please refer to~\cite{van2018hyperparameter} wherein the authors perform a detailed analysis of the important hyper-parameters (along with appropriate prior distributions for each) for a wide range of learning models.
\begin{figure*}[!ht]
    \begin{minipage}{0.28\textwidth}
        \centering
        \subcaption{Parzen Tree Estimator based\\ Hyperparameter Tuning}
        \begin{tabular}{p{1.3cm}p{0.4cm}p{0.4cm}p{0.4cm}p{0.4cm}}
            \toprule
            {} &  Prec. &  Rec. &  F1 &   Sup. \\
            \midrule
            $\neg$ Attrited &0.98& 0.98& 0.98&2543\\
            Attrited & 0.91& 0.88& 0.89&496\\
            acc.          & & & 0.97&3039\\
            macro avg.         & 0.94& 0.93& 0.94&3039\\
            wt. avg.      & 0.97& 0.97& 0.97&3039\\
        \bottomrule
        \end{tabular}
        \label{tab:hyp_tuning}
    \end{minipage}
    \hfill
    \begin{minipage}{0.28\textwidth}
        \centering
        \subcaption{No Hyperparameter Tuning\\ (Manually Set)}
        \begin{tabular}{p{1.3cm}p{0.4cm}p{0.4cm}p{0.4cm}p{0.4cm}}
            \toprule
            {} &  Prec. &  Rec. &  F1 &   Sup. \\
            \midrule
            $\neg$ Attrited & 0.97& 0.98& 0.97&2543 \\
            Attrited & 0.89& 0.84& 0.86&496 \\
            acc.          & & & 0.96&3039 \\
            macro avg.         & 0.93&0.91 & 0.92&3039 \\
            wt. avg.      & 0.96& 0.96& 0.96&3039 \\
            \bottomrule
        \end{tabular}
    \label{tab:no_hyp_tuning}
    \end{minipage}
    \hfill
    \begin{minipage}{0.42\textwidth}
    \centering
    \subcaption{Parameter Importance Plot}
    \includegraphics[scale=0.333]{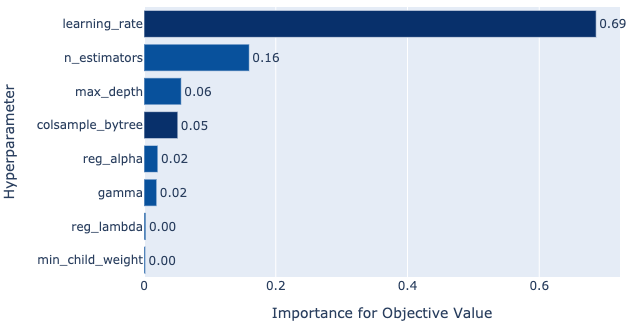}
    \label{fig:hyp_param_importance}
    \end{minipage}
    \caption{(a) Results of Attrition prediction task using XGboost classifier with hyperparameters tuned using tree structured Parzen estimator~\cite{bergstra2011algorithms,bergstra2013making}. (b) Results for the same task with XGBoost classifier trained with manually set hyperaparameters. We notice a drop in both precision (Prec.) and recall (Rec.) of the attrited customer (i.e., minority) class. (c) The parameter importance plot depicts importance of each hyper-parameter in trained classifier; we notice learning-rate is by far the most important hyperparameter to be tuned followed by n-estimators (i.e., number of trees), used in the XGboost ensemble. }
    \label{fig:hyp_optimization_plots}
\end{figure*}

The resurgent and recently popular learning methodology employing deep neural networks also has hyper-parameters like the hidden size of intermediate layers, the types of units to employ in the network architecture (fully connected, recurrent, convolutional), the types of activation functions (TanH, ReLU, Sigmoid), and types of regularizations to employ (Dropout layers, Batch Normalization, Strided-Convolutions, Pooling, $L_k$-norm regularization terms). In the context of deep learning models, this area of research is termed neural architecture search~\cite{elsken2019neural}.
Hyper-parameter optimization, has been conducted in multiple ways, however thus far a combination of manual tuning of hyper-parameters with grid-search approaches have proven to be the most effective~\cite{lecun2012efficient,larochelle2007empirical,hinton2012practical} in searching over the space of hyper-parameters. In~\cite{bergstra2012random}, the authors propose that random search (within a manually assigned range) of the hyper-parameter space yields a computationally cheap and an equivalent if not superior alternative to grid search based hyper-parameter optimization. Yet other approaches pose the hyper-parameter search as a Bayesian optimization problem~\cite{joy2016hyperparameter,snoek2012practical} over the search space.
Fig.~\ref{fig:hyp_optimization_plots} characterizes the optimization process on the learning task of detecting ``churn" or customer attrition using their activity patterns in the context of banking transactions. The figures therein yield an analysis of the hyper-parameter optimization process characterizing the relative importance of each hyper-parameter employed in the learning pipeline. As hyper-parameters play such a crucial role in learning (e.g., we notice from the statistics in Fig.~\ref{tab:hyp_tuning} that an XGBoost model is able to achieve a 3.5\% improvement in the F1 score of detecting attrited customers relative to an XGBoost variant without hyperparameter tuning i.e.,  Fig.~\ref{tab:no_hyp_tuning}), it is imperative that the part of a learning pipeline concerned with hyper-parameter optimization be explicitly and painstakingly documented so as to be reproducible and easily adaptable.
\subsection{`PEST' AntiPattern}\label{sec:baselines}
Like many applied scientific disciplines, machine learning (ML) research is driven by the empirical verification and validation of theoretical proposals. Novel contributions to applied machine learning research comprise (i) validation of previously unverified theoretical proposals, (ii) new theoretical proposals coupled with empirical verification, or (iii) effective augmentations to existing learning pipelines to yield improved empirical performance. Sound empirical verification requires a fair evaluation of the proposed approach w.r.t previously proposed approaches to assess
empirical performance. However, it is quite often the case that empirical verification of newly proposed ML methodologies is insufficient, flawed, or found wanting. In such cases, the reported empirical gains are actually just an occurrence of the \emph{Perceived Empirical SuperioriTy} (PEST) antipattern. 

For example, in~\cite{henderson2018deep}, the authors question claimed advances in reinforcement learning research due to the lack of significance metrics and variability of results. In~\cite{melis2017state}, the authors state that many years of claimed superiority in empirical performance in the field of language modeling is actually faulty and showcase that the well-known stacked LSTM architecture (with appropriate hyperparameter tuning) outperforms other more recent and more sophisticated architectures. In~\cite{mukhoti2018importance}, the authors highlight a flaw in many previous research works (in the context of Bayesian deep learning) wherein a well established baseline (Monte Carlo dropout) when run to completion (i.e., when learning is not cut-off preemptively by setting it to terminate after a specified number of iterations), achieves similar or superior results compared to the very same models which showcased superior results when introduced. The authors thereby motivate the need for identical experimental pipelines for comparison and evaluation of ML models. In~\cite{shen2018baseline}, authors conduct an extensive comparative analysis of the supposed state-of-the-art word embedding models with a
`simple-word-embedding-model'
(SWEM) and find that the SWEM model yields performance comparable or superior to previously claimed (and more complicated) state-of-the-art models. In our financial analytics context, we have found the KISS principle to encourage developers to try simple models first and to conduct an exhaustive comparison of models before advocating for specific methodologies. Recent benchmark pipelines like the GLUE and SQuAD benchmarks~\cite{wang2018glue,rajpurkar2016squad} are potential ways to address the PEST antipattern.
\subsection{Bad Credit Assignment AntiPattern}\label{sec:bad_credit_assignment}
Another frequent troubling trend in ML modeling is the failure to appropriately identify the source of performance gains in a modeling pipeline. As the peer-review process encourages technical novelty, quite often, research work focuses on proposing empirically superior, and complicated model architectures. Such empirical superiority is explained to be a function of the novel architecture while it is most often the case that the performance gains are in fact a function of clever problem formulations, data preprocessing, hyper-parameter tuning, or the application of existing well-established methods to interesting new tasks as detailed by~\cite{lipton2018troubling}. 

Whenever possible, it is imperative that effective ablation studies highlighting the performance gains of each component of a newly proposed learning models be included as part of the empirical evaluation. There must also be a concerted effort to train and evaluate baselines and the proposed model(s) in comparable experimental settings. Finally as noted in~\cite{lipton2018troubling}, if ablation studies are infeasible, quantifying the error behavior~\cite{kwiatkowski2013scaling} and robustness~\cite{cotterell2018all} of the proposed model can also yield significant insights about model behavior.
\subsection{Grade-your-own-Exam AntiPattern}\label{sec:testing_and_evaluation}
Usually modeling projects begin as curiosity-driven iterations to explore for potential traction. The measure of traction is calculated somewhat informally without formal 3rd-party review or validation. While not a problem at first, if the data science team continues this practice long enough, while building confidence in their results, they never validate them and cannot compare unvalidated results against other methods. To avoid this antipattern,
testing and evaluation data should be sampled independently, and for a robust performance analysis, should be kept hidden until model development is complete and must be used only for final valuation. 
In practice, it is not
uncommon for
model developers to have access to the final test set and by repeated testing against this known test set, modify their model accordingly to improve performance on the known test set. This practice called HARKing (Hypothesizing After Results are Known) has been detailed by Gencoglu et al.~\cite{gencoglu2019hark}. This leads to implicit data leakage. Cawley et. al.~\cite{cawley} discusses the potential effects of not having a statistically `pure' test set such as over-fitting and selection bias in performance evaluation.

The refactored solution here is not simple, but is essential and necessary for effective governance and oversight. Data science
teams must establish an independent `Ground Truth system' with APIs to receive and catalog all forecasts and the data that were used to make them. This system can provide a reliable date stamp that accurately reflects when any data object or forecast was actually made available or made and can help
track independent 3rd party metrics that will stand up to audit.

\subsection{Set \& Forget AntiPattern}\label{sec:concept_drift}
    A core assumption of machine learning (ML) pipelines is that the data generating process being sampled from (for training and when the model is deployed in production) generates samples that are \textit{independent} and \textit{identically distributed} (i.i.d). ML pipelines predominantly adopt a `set \& forget' mentality to model training and inference. However, it is quite often the case that the statistical properties of the target variable that the learning model is trying to predict change over time (\textit{concept drift}~\cite{widmer1996learning}). Decision support systems governed by data-driven models are required to effectively handle concept drift and yield accurate decisions. The primary technique to handle concept drift is learning rule sets using techniques based on decision trees and other similar interpretable tree-based approaches. Domingos et al.~\cite{domingos2000mining} proposed a model based on Hoeffding trees. Klienberg et al.~\cite{klinkenberg2004learning}, propose sliding window and instance weighting methods to maintain the learning model consistent with the most recent (albeit drifted) data. Various other approaches based on rule sets, Bayesian modeling have been developed for detection and adaptation of concept drift, details can be found in~\cite{gama2014survey,lu2018learning,webb2016characterizing}.
    An example of model drift adaptation can be seen in Chakraborty et al.~\cite{hqcd} for forecasting protest events. This work provides a use case wherein changes in surrogates can be used to detect change points in the target series with lower delay than just using the target's history.

\subsection{`Communicate with Ambivalence' AntiPattern}\label{sec:calibration}
Most ML pipelines are
tuned to generate
predictions but little
attention is paid to ensure that the model can sufficiently communicate information about its own uncertainty. A well-calibrated model is one where the Brier score (or similar) is carefully calibrated in its confidence. 
%See Fig.~\ref{briers} for examples of well- and poorly-calibrated models.
When poorly calibrated models
are placed in production,
it becomes difficult to
introduce compensatory
or remedial pipelines when
something goes wrong. 
Characterizing
model uncertainty is thus
a paramount feature for
large-scale deployment.
Recent work~\cite{bhatt2020uncertainty} shows that
in addition to explainability, conveying uncertainty can be a significant contributor to ensuring trust in ML pipelines.

\subsection{`Data Crisis as a Service' 
AntiPattern}\label{sec:data_crisis_as_a_service}
The development of models using data manually extracted and hygiened without recording the extraction or hygiene steps leads to
a massive 
data preparation challenge
for
later attempts to validate (or even deploy) ML
models.
 This is often the result of `sensitive' data that is selectively sanitized for the modelers by some third-party data steward organization that cannot adequately determine the risk associated with direct data access. The data preparation steps are effectively swept under the carpet and must be completely re-invented later, often with surprising impact on the models because the pipeline ends up producing different data.

The refactored solution here is to: (i) ensure that your enterprise sets up a professional data engineering practice that can quickly build and support new data pipelines that are governed and resilient; (ii) use
assertions to track
data as they move through
the pipeline, and
(iii) track the pedigree and lineage of all
data products, 
including intermediaries.
We've found graph databases to be ideal for maintaining linkages between data objects and the many assertions you must track.
\begin{figure*}[!ht]
    \begin{minipage}{0.45\textwidth}
    \centering
    \includegraphics[scale=0.34]{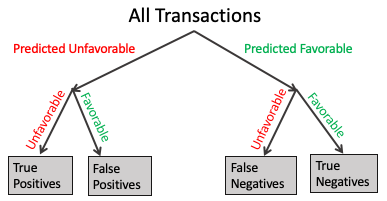}
    \subcaption{Current state of ML deployment pipeline evaluation focuses only on the single stage performance of the ML model.}
    \label{fig:ml_deployment_1}
    \end{minipage}
    \hfill
    \begin{minipage}{0.45\textwidth}
    \centering
    \includegraphics[scale=0.36]{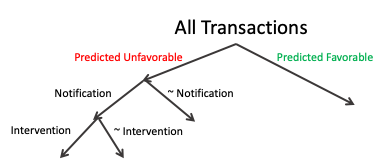}
    \subcaption{ML deployment pipelines are actually multi-stage decision systems with a hierarchical setup of learning, notification and intervention layers, each requiring evaluation.}
    \label{fig:ml_deployment_2}
    \end{minipage}
    \vfill
    \begin{minipage}{0.6\textwidth}
    \centering
    \includegraphics[scale=0.4]{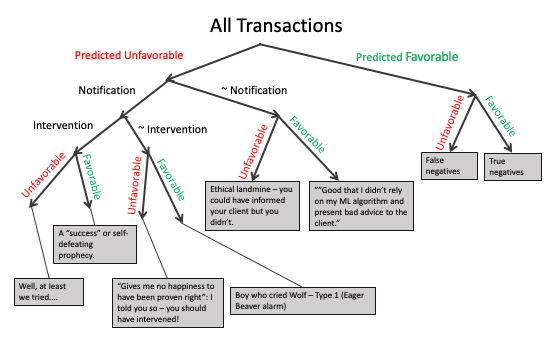}
    \subcaption{A characterization (with state-specific illustrations) of the various operational states the ML deployment pipeline can assume.}
    \label{fig:ml_deployment_3}
    \end{minipage}
    \caption{Characterization of the updated ML deployment pipeline and the need for monitoring an expanded set of possible operational states.}
\end{figure*}
\section{Rethinking ML Deployment}\label{sec:rethinkingmodeldeployment}
\iffalse 
\begin{figure*}[!ht]
    \begin{minipage}{0.45\textwidth}
    \centering
    \includegraphics[scale=0.38]{figures/modelops_stage_1.png}
    \subcaption{Current state of ML deployment pipeline evaluation focuses only on the single stage performance of the ML model.}
    \label{fig:ml_deployment_1}
    \end{minipage}
    \hfill
    \begin{minipage}{0.45\textwidth}
    \centering
    \includegraphics[scale=0.38]{figures/modelops_stage_2.png}
    \subcaption{ML deployment pipelines are actually multi-stage decision systems with a hierarchical setup of learning, notification and intervention layers, each requiring evaluation.}
    \label{fig:ml_deployment_2}
    \end{minipage}
    \vfill
    \begin{minipage}{0.6\textwidth}
    \centering
    \includegraphics[scale=0.4]{figures/modelops_stage_3.png}
    \subcaption{A characterization (with state-specific illustrations) of the various operational states the ML deployment pipeline can assume.}
    \label{fig:ml_deployment_3}
    \end{minipage}
    \caption{Characterization of the updated ML deployment pipeline and the need for monitoring an expanded set of possible operational states.}
\end{figure*}
\fi 
Machine Learning (ML) models are usually evaluated with metrics (e.g., precision, recall, confusion matrices serve as evaluation metrics in a classification setting) that are solely focused on characterizing the performance of the core learning model. However, production systems are often decision guidance systems with multiple additional notification (e.g., a process that raises an alert when the core learning model yields a particular prediction) and intervention (e.g., a process that carries out an appropriate action based on the results of the notification system) layers built on top of the core learning layer. 

Fig.~\ref{fig:ml_deployment_1} showcases the outcomes in a traditional (ML focused) evaluation pipeline wherein an ML model predicts a transaction to have a \emph{Favorable} or \emph{Unfavorable} outcome. Depending on the application, the definition of what is considered \emph{Favorable} or \emph{Unfavorable} may differ. 

For illustration, let us consider a fraud detection application, wherein an unfavorable outcome would be defined as a fraudulent transaction while legitimate transactions would be considered favorable outcomes. An ML model tasked with detecting fraudulent transactions would predict whether each transaction was \emph{Favorable} or \emph{Unfavorable}. In this context, Fig.~\ref{fig:ml_deployment_1} indicates that the deployed ML model pipeline may enter four possible states during its operation life-cycle. However, Fig.~\ref{fig:ml_deployment_2} showcases a slightly more realistic ML pipeline wherein notification (send alerts) and intervention (take appropriate action) layers are added on top of the ML model decisions to appropriately raise alerts or intervene to arrest progress of a potentially fraudulent transaction detected by the ML model.

The addition of these alerting and notification mechanisms which are imperative and ubiquitous in enterprise ML settings augment the number of possible states the ML pipeline may enter during its operation. These new states the model may enter create more nuanced situations with new dilemmas which are not highlighted by a simplistic evaluation approach like the one indicated in Fig.~\ref{fig:ml_deployment_1}. For example, if we observe the state the pipeline reaches if the ML model predicts a transaction to be fraudulent (i.e., \emph{unfavorable}) and the notification pipeline does not notify the client of the model decision, then if the transaction is actually fraudulent, then this situation is fraught with ethical ramifications. This exhaustive state representation of the ML decision pipeline in Fig.~\ref{fig:ml_deployment_3} allows us to explicitly add high penalties to such states allowing the ML, notification and intervention models to be trained cognizant of such penalties, essentially allowing fine-grained control of the learning and decision process. A more rigorous approach is to use a reinforcement learning formulation to track decision making and actions as models are put in production.
\section{Recommendations}
How do we make use of these lessons learnt and operationalize them in a production financial ML setting?
Specific recommendations we have include:
\begin{enumerate}
    \item Use AntiPatterns presented here to document a model management process to avoid costly but routine mistakes in model development, deployment, and approval.
    \item Use assertions to track data quality across the enterprise. This is crucial since ML models can be so dependent on faulty or noisy data and suitable checks and balances can ensure a safe operating environment for ML algorithms.
    \item Document data lineage along with transformations to support creation of `audit trails' so models can be situated back in time and in specific data slices for re-training or re-tuning.
    \item Use ensembles to maintain a palette of models including remedial and compensatory pipelines in the event of errors. Track model histories through the lifecycle of an application.
    \item Ensure human-in-the-loop operational capability at multiple levels. Use our model presented for rethinking ML deployment from Section~\ref{sec:rethinkingmodeldeployment} as a basis to support interventions and communication opportunities.
\end{enumerate}

Overall, the model development and management pipeline in our organization supports four classes of stakeholders: (i) the data steward (who holds custody of datasets and sets performance standards), (ii) the model developer (an ML person who designs algorithms), (iii) the model engineer (who places models in production and tracks performance), and (iv) the model certification authority (group of professionals who ensure compliance with standards and risk levels).
In particular, as ML models continue to make their way into more financial decision making systems, the model certification authority within the organization is crucial to ensuring regulatory compliance, from performance, safety, and auditability perspectives.
Bringing such multiple stakeholder groups together ensures a structured process where benefits and risks of ML models are well documented and understood at all stages of development and deployment.

{\small
\subsection*{Disclaimer:} BNY Mellon is the corporate brand of The Bank of New York Mellon Corporation and may be used to reference the corporation as a whole and/or its various subsidiaries generally.  This material does not constitute a recommendation by BNY Mellon of any kind.  The information herein is not intended to provide tax, legal, investment, accounting, financial or other professional advice on any matter, and should not be used or relied upon as such.  The views expressed within this material are those of the contributors and not necessarily those of BNY Mellon.  BNY Mellon has not independently verified the information contained in this material and makes no representation as to the accuracy, completeness, timeliness, merchantability or fitness for a specific purpose of the information provided in this material.  BNY Mellon assumes no direct or consequential liability for any errors in or reliance upon this material.}

\bibliographystyle{ACM-Reference-Format}
\bibliography{main}


\begin{thebibliography}{40}

%%% the Web address.

\ifx \showCODEN    \undefined \def \showCODEN     #1{\unskip}     \fi
\ifx \showDOI      \undefined \def \showDOI       #1{#1}\fi
\ifx \showISBNx    \undefined \def \showISBNx     #1{\unskip}     \fi
\ifx \showISBNxiii \undefined \def \showISBNxiii  #1{\unskip}     \fi
\ifx \showISSN     \undefined \def \showISSN      #1{\unskip}     \fi
\ifx \showLCCN     \undefined \def \showLCCN      #1{\unskip}     \fi
\ifx \shownote     \undefined \def \shownote      #1{#1}          \fi
\ifx \showarticletitle \undefined \def \showarticletitle #1{#1}   \fi
\ifx \showURL      \undefined \def \showURL       {\relax}        \fi
% invisible to TeX
\providecommand\bibfield[2]{#2}
\providecommand\bibinfo[2]{#2}
\providecommand\natexlab[1]{#1}
\providecommand\showeprint[2][]{arXiv:#2}

\bibitem[\protect\citeauthoryear{Bergstra, Bardenet, Bengio, and
  K{\'e}gl}{Bergstra et~al\mbox{.}}{2011}]%
        {bergstra2011algorithms}
\bibfield{author}{\bibinfo{person}{James Bergstra}, \bibinfo{person}{R{\'e}mi
  Bardenet}, \bibinfo{person}{Yoshua Bengio}, {and} \bibinfo{person}{Bal{\'a}zs
  K{\'e}gl}.} \bibinfo{year}{2011}\natexlab{}.
\newblock \showarticletitle{Algorithms for hyper-parameter optimization}. In
  \bibinfo{booktitle}{\emph{NeurIPS}}, Vol.~\bibinfo{volume}{24}.
\newblock


\bibitem[\protect\citeauthoryear{Bergstra and Bengio}{Bergstra and
  Bengio}{2012}]%
        {bergstra2012random}
\bibfield{author}{\bibinfo{person}{James Bergstra} {and}
  \bibinfo{person}{Yoshua Bengio}.} \bibinfo{year}{2012}\natexlab{}.
\newblock \showarticletitle{Random search for hyper-parameter optimization.}
\newblock \bibinfo{journal}{\emph{JMLR}} \bibinfo{volume}{13},
  \bibinfo{number}{2} (\bibinfo{year}{2012}).
\newblock


\bibitem[\protect\citeauthoryear{Bergstra, Yamins, and Cox}{Bergstra
  et~al\mbox{.}}{2013}]%
        {bergstra2013making}
\bibfield{author}{\bibinfo{person}{James Bergstra}, \bibinfo{person}{Daniel
  Yamins}, {and} \bibinfo{person}{David Cox}.} \bibinfo{year}{2013}\natexlab{}.
\newblock \showarticletitle{Making a science of model search: Hyperparameter
  optimization in hundreds of dimensions for vision architectures}. In
  \bibinfo{booktitle}{\emph{ICML}}. PMLR, \bibinfo{pages}{115--123}.
\newblock


\bibitem[\protect\citeauthoryear{Bhatt, Antor{\'a}n, Zhang, Liao, Sattigeri,
  Fogliato, Melan{\c{c}}on, Krishnan, Stanley, Tickoo, et~al\mbox{.}}{Bhatt
  et~al\mbox{.}}{2020}]%
        {bhatt2020uncertainty}
\bibfield{author}{\bibinfo{person}{Umang Bhatt}, \bibinfo{person}{Javier
  Antor{\'a}n}, \bibinfo{person}{Yunfeng Zhang}, \bibinfo{person}{Q~Vera Liao},
  \bibinfo{person}{Prasanna Sattigeri}, \bibinfo{person}{Riccardo Fogliato},
  \bibinfo{person}{Gabrielle~Gauthier Melan{\c{c}}on},
  \bibinfo{person}{Ranganath Krishnan}, \bibinfo{person}{Jason Stanley},
  \bibinfo{person}{Omesh Tickoo}, {et~al\mbox{.}}}
  \bibinfo{year}{2020}\natexlab{}.
\newblock \showarticletitle{Uncertainty as a form of transparency: Measuring,
  communicating, and using uncertainty}.
\newblock \bibinfo{journal}{\emph{arXiv preprint arXiv:2011.07586}}
  (\bibinfo{year}{2020}).
\newblock


\bibitem[\protect\citeauthoryear{Brown, Malveau, Brown, and HW~III}{Brown
  et~al\mbox{.}}{1998}]%
        {antipatterns}
\bibfield{author}{\bibinfo{person}{WJ Brown}, \bibinfo{person}{RC Malveau},
  \bibinfo{person}{McCormick Brown}, {and} \bibinfo{person}{Mowbray HW~III}.}
  \bibinfo{year}{1998}\natexlab{}.
\newblock \bibinfo{title}{TJ, AntiPatterns: Refactoring Software,
  Architectures, and Projects in Crisis}.
\newblock
\newblock


\bibitem[\protect\citeauthoryear{Cawley and Talbot}{Cawley and Talbot}{2010}]%
        {cawley}
\bibfield{author}{\bibinfo{person}{Gavin~C. Cawley} {and}
  \bibinfo{person}{Nicola~L.C. Talbot}.} \bibinfo{year}{2010}\natexlab{}.
\newblock \showarticletitle{On Over-Fitting in Model Selection and Subsequent
  Selection Bias in Performance Evaluation}.
\newblock \bibinfo{journal}{\emph{JMLR}}  \bibinfo{volume}{11}
  (\bibinfo{date}{Aug.} \bibinfo{year}{2010}), \bibinfo{pages}{2079–2107}.
\newblock
\showISSN{1532-4435}


\bibitem[\protect\citeauthoryear{Chakraborty, Muthiah, Tandon, and
  Ramakrishnan}{Chakraborty et~al\mbox{.}}{2016}]%
        {hqcd}
\bibfield{author}{\bibinfo{person}{Prithwish Chakraborty},
  \bibinfo{person}{Sathappan Muthiah}, \bibinfo{person}{Ravi Tandon}, {and}
  \bibinfo{person}{Naren Ramakrishnan}.} \bibinfo{year}{2016}\natexlab{}.
\newblock \showarticletitle{Hierarchical Quickest Change Detection via
  Surrogates}.
\newblock \bibinfo{journal}{\emph{arXiv preprint arXiv:1603.09739}}
  (\bibinfo{year}{2016}).
\newblock


\bibitem[\protect\citeauthoryear{Chawla, Bowyer, Hall, and Kegelmeyer}{Chawla
  et~al\mbox{.}}{2002}]%
        {smote}
\bibfield{author}{\bibinfo{person}{Nitesh~V. Chawla}, \bibinfo{person}{Kevin~W.
  Bowyer}, \bibinfo{person}{Lawrence~O. Hall}, {and} \bibinfo{person}{W.~Philip
  Kegelmeyer}.} \bibinfo{year}{2002}\natexlab{}.
\newblock \showarticletitle{SMOTE: Synthetic Minority over-Sampling Technique}.
\newblock \bibinfo{journal}{\emph{J. Artif. Int. Res.}} \bibinfo{volume}{16},
  \bibinfo{number}{1} (\bibinfo{date}{June} \bibinfo{year}{2002}),
  \bibinfo{pages}{321–357}.
\newblock
\showISSN{1076-9757}


\bibitem[\protect\citeauthoryear{Cotterell, Mielke, Eisner, and
  Roark}{Cotterell et~al\mbox{.}}{2018}]%
        {cotterell2018all}
\bibfield{author}{\bibinfo{person}{Ryan Cotterell}, \bibinfo{person}{Sabrina~J
  Mielke}, \bibinfo{person}{Jason Eisner}, {and} \bibinfo{person}{Brian
  Roark}.} \bibinfo{year}{2018}\natexlab{}.
\newblock \showarticletitle{Are all languages equally hard to language-model?}
\newblock \bibinfo{journal}{\emph{arXiv preprint arXiv:1806.03743}}
  (\bibinfo{year}{2018}).
\newblock


\bibitem[\protect\citeauthoryear{Domingos and Hulten}{Domingos and
  Hulten}{2000}]%
        {domingos2000mining}
\bibfield{author}{\bibinfo{person}{Pedro Domingos} {and} \bibinfo{person}{Geoff
  Hulten}.} \bibinfo{year}{2000}\natexlab{}.
\newblock \showarticletitle{Mining high-speed data streams}. In
  \bibinfo{booktitle}{\emph{ACM SIGKDD}}. \bibinfo{pages}{71--80}.
\newblock


\bibitem[\protect\citeauthoryear{Elsken, Metzen, Hutter, et~al\mbox{.}}{Elsken
  et~al\mbox{.}}{2019}]%
        {elsken2019neural}
\bibfield{author}{\bibinfo{person}{Thomas Elsken}, \bibinfo{person}{Jan~Hendrik
  Metzen}, \bibinfo{person}{Frank Hutter}, {et~al\mbox{.}}}
  \bibinfo{year}{2019}\natexlab{}.
\newblock \showarticletitle{Neural architecture search: A survey.}
\newblock \bibinfo{journal}{\emph{JMLR}} \bibinfo{volume}{20},
  \bibinfo{number}{55} (\bibinfo{year}{2019}), \bibinfo{pages}{1--21}.
\newblock


\bibitem[\protect\citeauthoryear{Gama, {\v{Z}}liobait{\.e}, Bifet, Pechenizkiy,
  and Bouchachia}{Gama et~al\mbox{.}}{2014}]%
        {gama2014survey}
\bibfield{author}{\bibinfo{person}{Jo{\~a}o Gama}, \bibinfo{person}{Indr{\.e}
  {\v{Z}}liobait{\.e}}, \bibinfo{person}{Albert Bifet}, \bibinfo{person}{Mykola
  Pechenizkiy}, {and} \bibinfo{person}{Abdelhamid Bouchachia}.}
  \bibinfo{year}{2014}\natexlab{}.
\newblock \showarticletitle{A survey on concept drift adaptation}.
\newblock \bibinfo{journal}{\emph{ACM CSUR}} \bibinfo{volume}{46},
  \bibinfo{number}{4} (\bibinfo{year}{2014}), \bibinfo{pages}{1--37}.
\newblock


\bibitem[\protect\citeauthoryear{Gencoglu, van Gils, Guldogan, Morikawa,
  S{\"u}zen, Gruber, Leinonen, and Huttunen}{Gencoglu et~al\mbox{.}}{2019}]%
        {gencoglu2019hark}
\bibfield{author}{\bibinfo{person}{Oguzhan Gencoglu}, \bibinfo{person}{Mark van
  Gils}, \bibinfo{person}{Esin Guldogan}, \bibinfo{person}{Chamin Morikawa},
  \bibinfo{person}{Mehmet S{\"u}zen}, \bibinfo{person}{Mathias Gruber},
  \bibinfo{person}{Jussi Leinonen}, {and} \bibinfo{person}{Heikki Huttunen}.}
  \bibinfo{year}{2019}\natexlab{}.
\newblock \showarticletitle{HARK Side of Deep Learning--From Grad Student
  Descent to Automated Machine Learning}.
\newblock \bibinfo{journal}{\emph{arXiv preprint arXiv:1904.07633}}
  (\bibinfo{year}{2019}).
\newblock


\bibitem[\protect\citeauthoryear{Henderson, Islam, Bachman, Pineau, Precup, and
  Meger}{Henderson et~al\mbox{.}}{2018}]%
        {henderson2018deep}
\bibfield{author}{\bibinfo{person}{Peter Henderson}, \bibinfo{person}{Riashat
  Islam}, \bibinfo{person}{Philip Bachman}, \bibinfo{person}{Joelle Pineau},
  \bibinfo{person}{Doina Precup}, {and} \bibinfo{person}{David Meger}.}
  \bibinfo{year}{2018}\natexlab{}.
\newblock \showarticletitle{Deep reinforcement learning that matters}. In
  \bibinfo{booktitle}{\emph{Proceedings of the AAAI Conference on Artificial
  Intelligence}}, Vol.~\bibinfo{volume}{32}.
\newblock


\bibitem[\protect\citeauthoryear{Hinton}{Hinton}{2012}]%
        {hinton2012practical}
\bibfield{author}{\bibinfo{person}{Geoffrey~E Hinton}.}
  \bibinfo{year}{2012}\natexlab{}.
\newblock \showarticletitle{A practical guide to training restricted Boltzmann
  machines}.
\newblock In \bibinfo{booktitle}{\emph{Neural networks: Tricks of the trade}}.
  \bibinfo{publisher}{Springer}, \bibinfo{pages}{599--619}.
\newblock


\bibitem[\protect\citeauthoryear{Isbell}{Isbell}{2020}]%
        {isbell2020}
\bibfield{author}{\bibinfo{person}{Charles Isbell}.}
  \bibinfo{year}{2020}\natexlab{}.
\newblock \showarticletitle{You Can’t Escape Hyperparameters and Latent
  Variables: Machine Learning as a Software Engineering Enterpri}.
\newblock \bibinfo{journal}{\emph{NeurIPS 2020}} (\bibinfo{date}{Dec}
  \bibinfo{year}{2020}).
\newblock


\bibitem[\protect\citeauthoryear{Joy, Rana, Gupta, and Venkatesh}{Joy
  et~al\mbox{.}}{2016}]%
        {joy2016hyperparameter}
\bibfield{author}{\bibinfo{person}{Tinu~Theckel Joy}, \bibinfo{person}{Santu
  Rana}, \bibinfo{person}{Sunil Gupta}, {and} \bibinfo{person}{Svetha
  Venkatesh}.} \bibinfo{year}{2016}\natexlab{}.
\newblock \showarticletitle{Hyperparameter tuning for big data using Bayesian
  optimisation}. In \bibinfo{booktitle}{\emph{ICPR}}. IEEE,
  \bibinfo{pages}{2574--2579}.
\newblock


\bibitem[\protect\citeauthoryear{Klinkenberg}{Klinkenberg}{2004}]%
        {klinkenberg2004learning}
\bibfield{author}{\bibinfo{person}{Ralf Klinkenberg}.}
  \bibinfo{year}{2004}\natexlab{}.
\newblock \showarticletitle{Learning drifting concepts: Example selection vs.
  example weighting}.
\newblock \bibinfo{journal}{\emph{Intelligent data analysis}}
  \bibinfo{volume}{8}, \bibinfo{number}{3} (\bibinfo{year}{2004}),
  \bibinfo{pages}{281--300}.
\newblock


\bibitem[\protect\citeauthoryear{Kwiatkowski, Choi, Artzi, and
  Zettlemoyer}{Kwiatkowski et~al\mbox{.}}{2013}]%
        {kwiatkowski2013scaling}
\bibfield{author}{\bibinfo{person}{Tom Kwiatkowski}, \bibinfo{person}{Eunsol
  Choi}, \bibinfo{person}{Yoav Artzi}, {and} \bibinfo{person}{Luke
  Zettlemoyer}.} \bibinfo{year}{2013}\natexlab{}.
\newblock \showarticletitle{Scaling semantic parsers with on-the-fly ontology
  matching}. In \bibinfo{booktitle}{\emph{EMNLP}}. \bibinfo{pages}{1545--1556}.
\newblock


\bibitem[\protect\citeauthoryear{Larochelle, Erhan, Courville, Bergstra, and
  Bengio}{Larochelle et~al\mbox{.}}{2007}]%
        {larochelle2007empirical}
\bibfield{author}{\bibinfo{person}{Hugo Larochelle}, \bibinfo{person}{Dumitru
  Erhan}, \bibinfo{person}{Aaron Courville}, \bibinfo{person}{James Bergstra},
  {and} \bibinfo{person}{Yoshua Bengio}.} \bibinfo{year}{2007}\natexlab{}.
\newblock \showarticletitle{An empirical evaluation of deep architectures on
  problems with many factors of variation}. In
  \bibinfo{booktitle}{\emph{ICML}}. \bibinfo{pages}{473--480}.
\newblock


\bibitem[\protect\citeauthoryear{LeCun, Bottou, Orr, and M{\"u}ller}{LeCun
  et~al\mbox{.}}{2012}]%
        {lecun2012efficient}
\bibfield{author}{\bibinfo{person}{Yann~A LeCun}, \bibinfo{person}{L{\'e}on
  Bottou}, \bibinfo{person}{Genevieve~B Orr}, {and}
  \bibinfo{person}{Klaus-Robert M{\"u}ller}.} \bibinfo{year}{2012}\natexlab{}.
\newblock \showarticletitle{Efficient backprop}.
\newblock In \bibinfo{booktitle}{\emph{Neural networks: Tricks of the trade}}.
  \bibinfo{publisher}{Springer}, \bibinfo{pages}{9--48}.
\newblock


\bibitem[\protect\citeauthoryear{Lipton and Steinhardt}{Lipton and
  Steinhardt}{2018}]%
        {lipton2018troubling}
\bibfield{author}{\bibinfo{person}{Zachary~C Lipton} {and}
  \bibinfo{person}{Jacob Steinhardt}.} \bibinfo{year}{2018}\natexlab{}.
\newblock \showarticletitle{Troubling trends in machine learning scholarship}.
\newblock \bibinfo{journal}{\emph{arXiv preprint arXiv:1807.03341}}
  (\bibinfo{year}{2018}).
\newblock


\bibitem[\protect\citeauthoryear{Lu, Liu, Dong, Gu, Gama, and Zhang}{Lu
  et~al\mbox{.}}{2018}]%
        {lu2018learning}
\bibfield{author}{\bibinfo{person}{Jie Lu}, \bibinfo{person}{Anjin Liu},
  \bibinfo{person}{Fan Dong}, \bibinfo{person}{Feng Gu}, \bibinfo{person}{Joao
  Gama}, {and} \bibinfo{person}{Guangquan Zhang}.}
  \bibinfo{year}{2018}\natexlab{}.
\newblock \showarticletitle{Learning under concept drift: A review}.
\newblock \bibinfo{journal}{\emph{IEEE TKDE}} \bibinfo{volume}{31},
  \bibinfo{number}{12} (\bibinfo{year}{2018}), \bibinfo{pages}{2346--2363}.
\newblock


\bibitem[\protect\citeauthoryear{Melis, Dyer, and Blunsom}{Melis
  et~al\mbox{.}}{2017}]%
        {melis2017state}
\bibfield{author}{\bibinfo{person}{G{\'a}bor Melis}, \bibinfo{person}{Chris
  Dyer}, {and} \bibinfo{person}{Phil Blunsom}.}
  \bibinfo{year}{2017}\natexlab{}.
\newblock \showarticletitle{On the state of the art of evaluation in neural
  language models}.
\newblock \bibinfo{journal}{\emph{arXiv preprint arXiv:1707.05589}}
  (\bibinfo{year}{2017}).
\newblock


\bibitem[\protect\citeauthoryear{Mitchell, Wu, Zaldivar, Barnes, Vasserman,
  Hutchinson, Spitzer, Raji, and Gebru}{Mitchell et~al\mbox{.}}{2019}]%
        {modelcards}
\bibfield{author}{\bibinfo{person}{Margaret Mitchell}, \bibinfo{person}{Simone
  Wu}, \bibinfo{person}{Andrew Zaldivar}, \bibinfo{person}{Parker Barnes},
  \bibinfo{person}{Lucy Vasserman}, \bibinfo{person}{Ben Hutchinson},
  \bibinfo{person}{Elena Spitzer}, \bibinfo{person}{Inioluwa~Deborah Raji},
  {and} \bibinfo{person}{Timnit Gebru}.} \bibinfo{year}{2019}\natexlab{}.
\newblock \showarticletitle{Model cards for model reporting}. In
  \bibinfo{booktitle}{\emph{FAccT}}. \bibinfo{pages}{220--229}.
\newblock


\bibitem[\protect\citeauthoryear{Mukhoti, Stenetorp, and Gal}{Mukhoti
  et~al\mbox{.}}{2018}]%
        {mukhoti2018importance}
\bibfield{author}{\bibinfo{person}{Jishnu Mukhoti}, \bibinfo{person}{Pontus
  Stenetorp}, {and} \bibinfo{person}{Yarin Gal}.}
  \bibinfo{year}{2018}\natexlab{}.
\newblock \showarticletitle{On the importance of strong baselines in bayesian
  deep learning}.
\newblock \bibinfo{journal}{\emph{arXiv preprint arXiv:1811.09385}}
  (\bibinfo{year}{2018}).
\newblock


\bibitem[\protect\citeauthoryear{Nguyen, Schulze, and Osborne}{Nguyen
  et~al\mbox{.}}{2019}]%
        {nguyen2019bayesian}
\bibfield{author}{\bibinfo{person}{Vu Nguyen}, \bibinfo{person}{Sebastian
  Schulze}, {and} \bibinfo{person}{Michael~A Osborne}.}
  \bibinfo{year}{2019}\natexlab{}.
\newblock \showarticletitle{Bayesian optimization for iterative learning}.
\newblock \bibinfo{journal}{\emph{arXiv preprint arXiv:1909.09593}}
  (\bibinfo{year}{2019}).
\newblock


\bibitem[\protect\citeauthoryear{Paleyes, Urma, and Lawrence}{Paleyes
  et~al\mbox{.}}{2020}]%
        {paleyes2020challenges}
\bibfield{author}{\bibinfo{person}{Andrei Paleyes},
  \bibinfo{person}{Raoul-Gabriel Urma}, {and} \bibinfo{person}{Neil~D
  Lawrence}.} \bibinfo{year}{2020}\natexlab{}.
\newblock \showarticletitle{Challenges in deploying machine learning: a survey
  of case studies}.
\newblock \bibinfo{journal}{\emph{arXiv preprint arXiv:2011.09926}}
  (\bibinfo{year}{2020}).
\newblock


\bibitem[\protect\citeauthoryear{Probst, Boulesteix, and Bischl}{Probst
  et~al\mbox{.}}{2019}]%
        {probst2019tunability}
\bibfield{author}{\bibinfo{person}{Philipp Probst}, \bibinfo{person}{Anne-Laure
  Boulesteix}, {and} \bibinfo{person}{Bernd Bischl}.}
  \bibinfo{year}{2019}\natexlab{}.
\newblock \showarticletitle{Tunability: Importance of hyperparameters of
  machine learning algorithms.}
\newblock \bibinfo{journal}{\emph{JMLR}} \bibinfo{volume}{20},
  \bibinfo{number}{53} (\bibinfo{year}{2019}), \bibinfo{pages}{1--32}.
\newblock


\bibitem[\protect\citeauthoryear{Rajpurkar, Zhang, Lopyrev, and
  Liang}{Rajpurkar et~al\mbox{.}}{2016}]%
        {rajpurkar2016squad}
\bibfield{author}{\bibinfo{person}{Pranav Rajpurkar}, \bibinfo{person}{Jian
  Zhang}, \bibinfo{person}{Konstantin Lopyrev}, {and} \bibinfo{person}{Percy
  Liang}.} \bibinfo{year}{2016}\natexlab{}.
\newblock \showarticletitle{Squad: 100,000+ questions for machine comprehension
  of text}.
\newblock \bibinfo{journal}{\emph{arXiv preprint arXiv:1606.05250}}
  (\bibinfo{year}{2016}).
\newblock


\bibitem[\protect\citeauthoryear{Ramakrishnan, Butler, Muthiah,
  et~al\mbox{.}}{Ramakrishnan et~al\mbox{.}}{2014}]%
        {beatingthenews}
\bibfield{author}{\bibinfo{person}{Naren Ramakrishnan},
  \bibinfo{person}{Patrick Butler}, \bibinfo{person}{Sathappan Muthiah},
  {et~al\mbox{.}}} \bibinfo{year}{2014}\natexlab{}.
\newblock \showarticletitle{'Beating the News' with EMBERS: Forecasting Civil
  Unrest Using Open Source Indicators}. In \bibinfo{booktitle}{\emph{ACM
  SIGKDD}} \emph{(\bibinfo{series}{KDD '14})}. \bibinfo{publisher}{Association
  for Computing Machinery}.
\newblock


\bibitem[\protect\citeauthoryear{Ribeiro, Singh, and Guestrin}{Ribeiro
  et~al\mbox{.}}{2016}]%
        {ribeiro2016should}
\bibfield{author}{\bibinfo{person}{Marco~Tulio Ribeiro},
  \bibinfo{person}{Sameer Singh}, {and} \bibinfo{person}{Carlos Guestrin}.}
  \bibinfo{year}{2016}\natexlab{}.
\newblock \showarticletitle{" Why should i trust you?" Explaining the
  predictions of any classifier}. In \bibinfo{booktitle}{\emph{ACM SIGKDD}}.
  \bibinfo{pages}{1135--1144}.
\newblock


\bibitem[\protect\citeauthoryear{Samala, Chan, Hadjiiski, and Koneru}{Samala
  et~al\mbox{.}}{2020}]%
        {samala2020hazards}
\bibfield{author}{\bibinfo{person}{Ravi~K Samala}, \bibinfo{person}{Heang-Ping
  Chan}, \bibinfo{person}{Lubomir Hadjiiski}, {and} \bibinfo{person}{Sathvik
  Koneru}.} \bibinfo{year}{2020}\natexlab{}.
\newblock \showarticletitle{Hazards of data leakage in machine learning: a
  study on classification of breast cancer using deep neural networks}. In
  \bibinfo{booktitle}{\emph{Medical Imaging 2020: Computer-Aided Diagnosis}},
  Vol.~\bibinfo{volume}{11314}. International Society for Optics and Photonics,
  \bibinfo{pages}{1131416}.
\newblock


\bibitem[\protect\citeauthoryear{Sculley, Holt, Golovin, Davydov, Phillips,
  Ebner, Chaudhary, Young, Crespo, and Dennison}{Sculley et~al\mbox{.}}{2015}]%
        {sculley2015hidden}
\bibfield{author}{\bibinfo{person}{David Sculley}, \bibinfo{person}{Gary Holt},
  \bibinfo{person}{Daniel Golovin}, \bibinfo{person}{Eugene Davydov},
  \bibinfo{person}{Todd Phillips}, \bibinfo{person}{Dietmar Ebner},
  \bibinfo{person}{Vinay Chaudhary}, \bibinfo{person}{Michael Young},
  \bibinfo{person}{Jean-Francois Crespo}, {and} \bibinfo{person}{Dan
  Dennison}.} \bibinfo{year}{2015}\natexlab{}.
\newblock \showarticletitle{Hidden technical debt in machine learning systems}.
\newblock \bibinfo{journal}{\emph{NeurIPS}}  \bibinfo{volume}{28}
  (\bibinfo{year}{2015}), \bibinfo{pages}{2503--2511}.
\newblock


\bibitem[\protect\citeauthoryear{Shen, Wang, Wang, Min, Su, Zhang, Li, Henao,
  and Carin}{Shen et~al\mbox{.}}{2018}]%
        {shen2018baseline}
\bibfield{author}{\bibinfo{person}{Dinghan Shen}, \bibinfo{person}{Guoyin
  Wang}, \bibinfo{person}{Wenlin Wang}, \bibinfo{person}{Martin~Renqiang Min},
  \bibinfo{person}{Qinliang Su}, \bibinfo{person}{Yizhe Zhang},
  \bibinfo{person}{Chunyuan Li}, \bibinfo{person}{Ricardo Henao}, {and}
  \bibinfo{person}{Lawrence Carin}.} \bibinfo{year}{2018}\natexlab{}.
\newblock \showarticletitle{Baseline needs more love: On simple
  word-embedding-based models and associated pooling mechanisms}.
\newblock \bibinfo{journal}{\emph{arXiv preprint arXiv:1805.09843}}
  (\bibinfo{year}{2018}).
\newblock


\bibitem[\protect\citeauthoryear{Snoek, Larochelle, and Adams}{Snoek
  et~al\mbox{.}}{2012}]%
        {snoek2012practical}
\bibfield{author}{\bibinfo{person}{Jasper Snoek}, \bibinfo{person}{Hugo
  Larochelle}, {and} \bibinfo{person}{Ryan~P Adams}.}
  \bibinfo{year}{2012}\natexlab{}.
\newblock \showarticletitle{Practical bayesian optimization of machine learning
  algorithms}.
\newblock \bibinfo{journal}{\emph{arXiv preprint arXiv:1206.2944}}
  (\bibinfo{year}{2012}).
\newblock


\bibitem[\protect\citeauthoryear{Van~Rijn and Hutter}{Van~Rijn and
  Hutter}{2018}]%
        {van2018hyperparameter}
\bibfield{author}{\bibinfo{person}{Jan~N Van~Rijn} {and} \bibinfo{person}{Frank
  Hutter}.} \bibinfo{year}{2018}\natexlab{}.
\newblock \showarticletitle{Hyperparameter importance across datasets}. In
  \bibinfo{booktitle}{\emph{ACM SIGKDD}}. \bibinfo{pages}{2367--2376}.
\newblock


\bibitem[\protect\citeauthoryear{Wang, Singh, Michael, Hill, Levy, and
  Bowman}{Wang et~al\mbox{.}}{2018}]%
        {wang2018glue}
\bibfield{author}{\bibinfo{person}{Alex Wang}, \bibinfo{person}{Amanpreet
  Singh}, \bibinfo{person}{Julian Michael}, \bibinfo{person}{Felix Hill},
  \bibinfo{person}{Omer Levy}, {and} \bibinfo{person}{Samuel~R Bowman}.}
  \bibinfo{year}{2018}\natexlab{}.
\newblock \showarticletitle{GLUE: A multi-task benchmark and analysis platform
  for natural language understanding}.
\newblock \bibinfo{journal}{\emph{arXiv preprint arXiv:1804.07461}}
  (\bibinfo{year}{2018}).
\newblock


\bibitem[\protect\citeauthoryear{Webb, Hyde, Cao, Nguyen, and Petitjean}{Webb
  et~al\mbox{.}}{2016}]%
        {webb2016characterizing}
\bibfield{author}{\bibinfo{person}{Geoffrey~I Webb}, \bibinfo{person}{Roy
  Hyde}, \bibinfo{person}{Hong Cao}, \bibinfo{person}{Hai~Long Nguyen}, {and}
  \bibinfo{person}{Francois Petitjean}.} \bibinfo{year}{2016}\natexlab{}.
\newblock \showarticletitle{Characterizing concept drift}.
\newblock \bibinfo{journal}{\emph{Data Mining and Knowledge Discovery}}
  \bibinfo{volume}{30}, \bibinfo{number}{4} (\bibinfo{year}{2016}),
  \bibinfo{pages}{964--994}.
\newblock


\bibitem[\protect\citeauthoryear{Widmer and Kubat}{Widmer and Kubat}{1996}]%
        {widmer1996learning}
\bibfield{author}{\bibinfo{person}{Gerhard Widmer} {and}
  \bibinfo{person}{Miroslav Kubat}.} \bibinfo{year}{1996}\natexlab{}.
\newblock \showarticletitle{Learning in the presence of concept drift and
  hidden contexts}.
\newblock \bibinfo{journal}{\emph{Machine learning}} \bibinfo{volume}{23},
  \bibinfo{number}{1} (\bibinfo{year}{1996}), \bibinfo{pages}{69--101}.
\newblock


\end{thebibliography}

% \appendix









\end{document}